\pdfoutput=1

\def\mathbi#1{\textbf{\em #1}}

\documentclass[11pt]{article}
\usepackage{enumitem}
\usepackage{times}
\usepackage{latexsym}
\usepackage{paralist}
\usepackage[T1]{fontenc}

\usepackage[utf8]{inputenc}
\usepackage{color}
\usepackage{graphicx}
\usepackage{threeparttable}
\usepackage[]{acl}
\usepackage{microtype}
\usepackage{amsmath}
\usepackage{amssymb}
\usepackage{latexsym}
\usepackage{times}
\usepackage{soul}
\usepackage{url}

\usepackage{graphicx}
\usepackage{amsthm}
\usepackage{amssymb}
\usepackage{amsmath}
\usepackage{multirow}
\usepackage{color}
\usepackage{booktabs}
\usepackage{algorithm}
\usepackage{algorithmic}
\usepackage{newfloat}
\usepackage{listings}
\usepackage{helvet}  
\usepackage{courier}  
\usepackage{graphicx} 
\usepackage{verbatim}
\usepackage{subfigure}
\usepackage{svg}
\usepackage{bm}
\usepackage{pifont}

\usepackage[T1]{fontenc}

\usepackage[utf8]{inputenc}

\usepackage{microtype}

\title{Few-shot Named Entity Recognition with Entity-level Prototypical Network Enhanced by Dispersedly Distributed  Prototypes}
\author{Bin Ji, Shasha Li, Shaoduo Gan, Jie Yu, Jun Ma, Huijun Liu \\
         College of Computer, National University of Defense Technology}

\begin{document}
\maketitle
\begin{abstract}
Few-shot named entity recognition (NER) enables us to build a NER system for a new domain using very few labeled examples.
However, existing prototypical networks for this task suffer from roughly estimated label dependency and closely distributed prototypes, thus often causing misclassifications. 
To address the above issues, we propose \textbf{EP-Net}, an \textbf{E}ntity-level \textbf{P}rototypical \textbf{Net}work enhanced by dispersedly distributed prototypes. EP-Net builds entity-level prototypes and considers text spans to be candidate entities, so it no longer requires the label dependency.
In addition, {EP-Net} trains the prototypes from scratch to distribute them dispersedly and aligns spans to prototypes in the embedding space using a space projection.
Experimental results on two evaluation tasks and the Few-NERD settings demonstrate that EP-Net consistently outperforms the previous strong models in terms of overall performance. Extensive analyses further validate the effectiveness of EP-Net.
\end{abstract}

\section{Introduction}
As a core language understanding task, named entity recognition (NER) faces rapid domain shifting. When transferring NER systems to new domains, one of the primary challenges is dealing with the mismatch of entity types \cite{yang}. For example, only 2 types are overlapped between I2B2 \cite{i2b2} and OntoNotes  \cite{ontonotes}{, which have} 23 and 18 entity types, respectively.
Unfortunately, annotating a new domain takes considerable time and effort{s}, let alone {the} domain knowledge {required} \cite{hou}.
Few-shot NER is targeted in this scenario since it can transfer prior experience from resource-rich (source) domains to resource-scarce (target) domains.

\begin{figure}[t]
\centering
\includegraphics[width=0.48\textwidth]{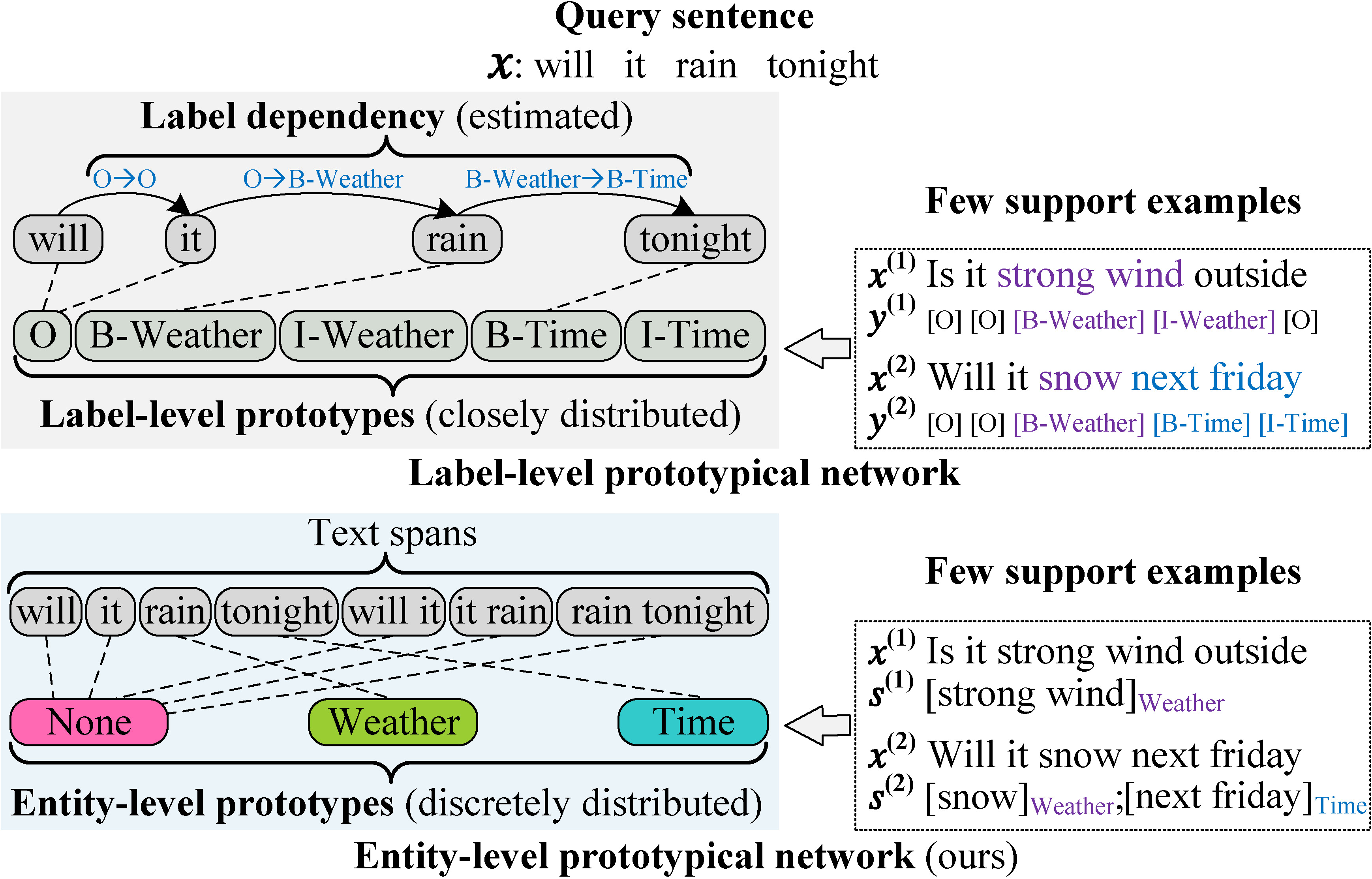} 
\caption{A comparison of token- and entity-level prototypical networks for few-shot NER, where the former builds prototypes for token labels and requires label dependency, while the latter builds {prototypes} for entity types and does not require label dependency. The dotted line denotes that the pair of token-prototype (or span-prototype) is the most similar. For clarity, we only list spans with lengths less than 2 and assume there are only 2 pre-defined entity types.}\label{figure1}
\label{model2}
\end{figure}

\begin{figure}[] 
\centering
\subfigure[Copying method]{
\begin{minipage}[t]{0.22\textwidth}
\centering
\includegraphics[width=1.51in]{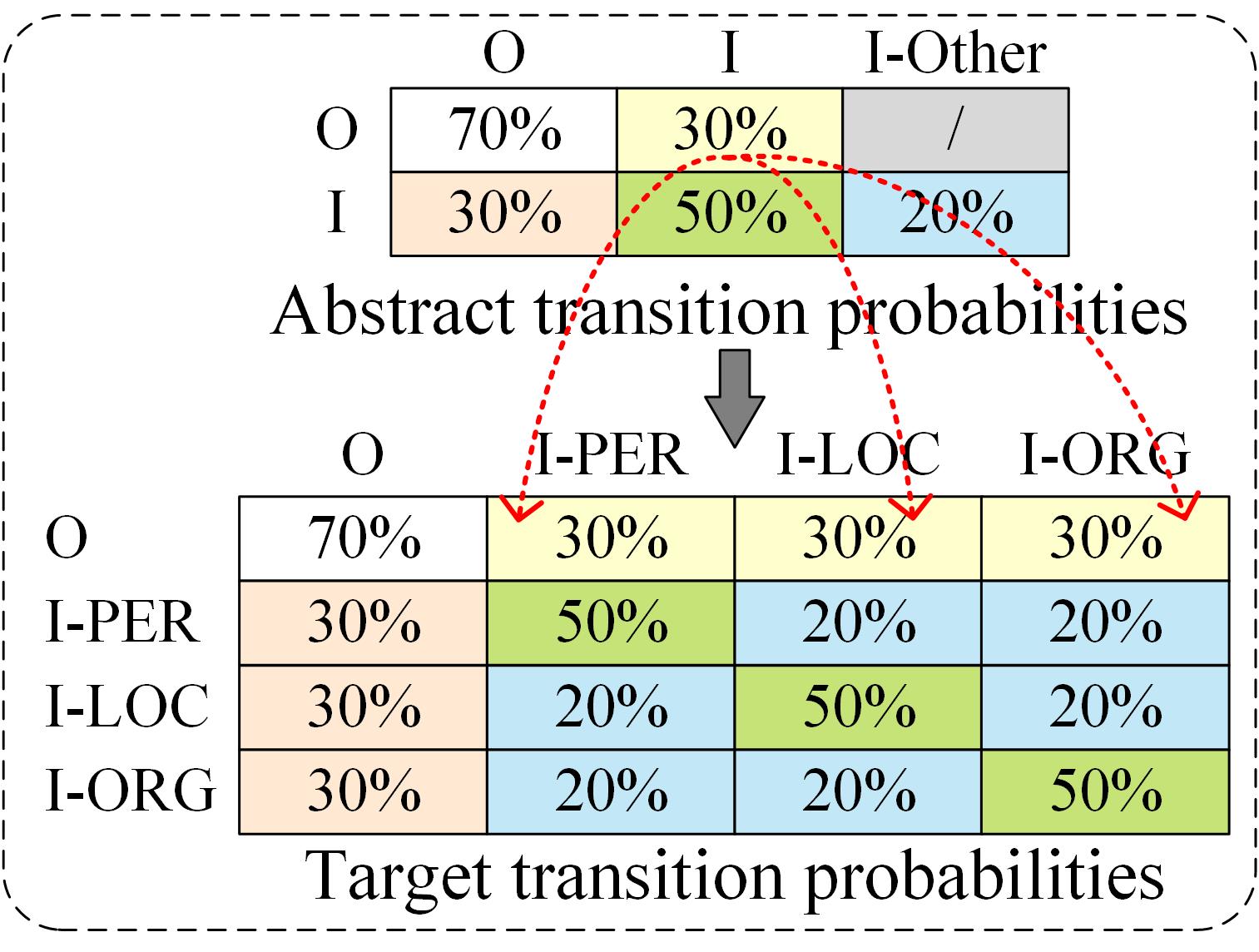}
\end{minipage}%
}%
\subfigure[Even distribution method]{
\begin{minipage}[t]{0.27\textwidth}
\centering
\includegraphics[width=1.51in]{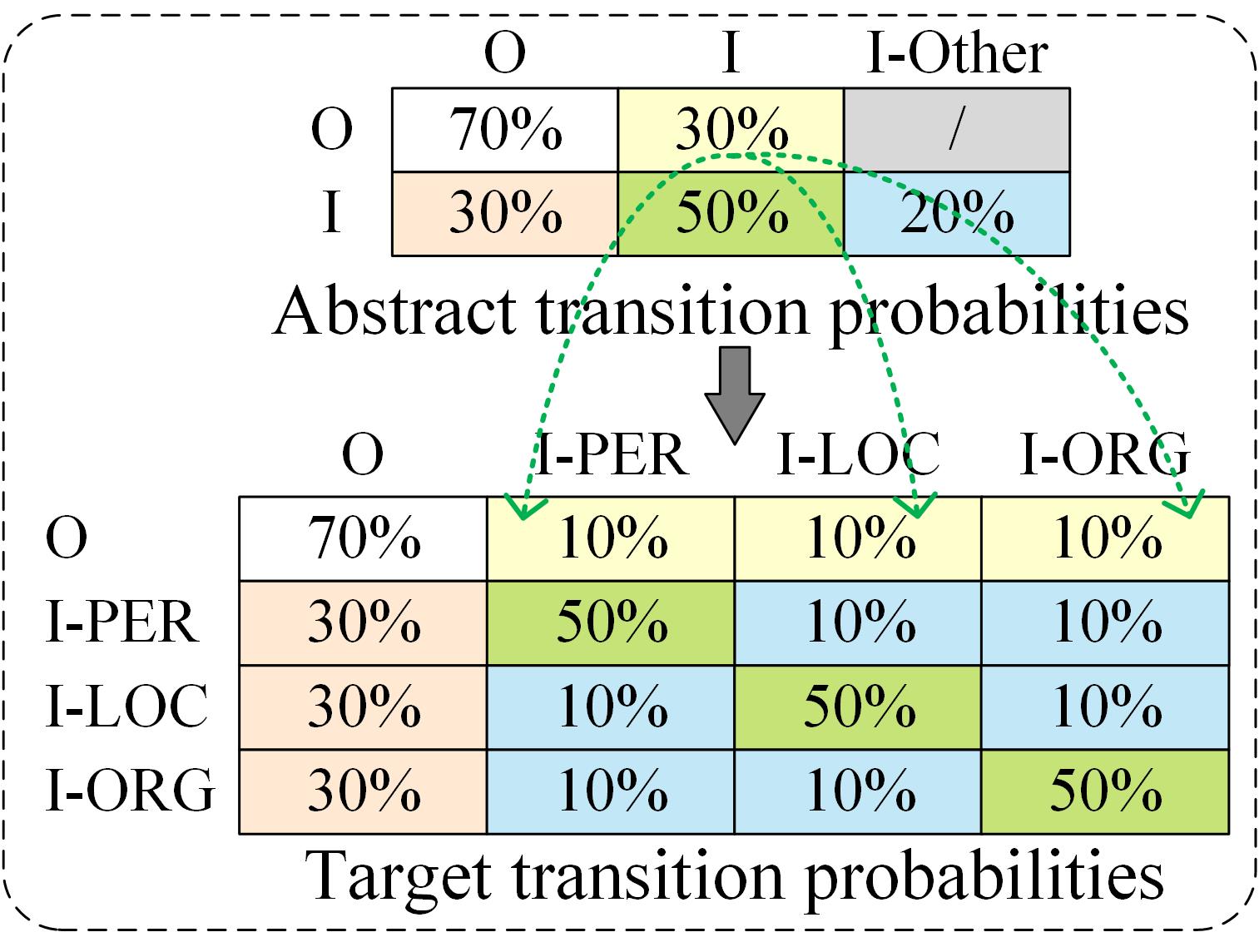}
\end{minipage}%
}%
\centering
\caption{Methods of transferring estimated label dependency. In each method, the abstracts and corresponding targets are displayed in the same color. Method (a) copies the abstracts to the targets, whereas method (b) distributes them equally to the targets.}\label{figure2}
\end{figure}

Previous few-shot NER models \cite{fri,hou,yang,tong} generally formulate the task as a sequence labeling task and employ token-level prototypical networks \cite{prototypical}. These models first obtain token labels according to the {most similar} token-prototype pair and then obtain entities based on these labels, as Figure \ref{figure1} shows.
The sequence labeling benefits from label dependency \cite{hou}.
{However, when it comes to few-shot NER models, the label dependency is off the table, because a few labeled data is way insufficient to learn the reliable dependency, and the label sets could vary from domain to domain.
To tackle this, some methods try to transfer roughly estimated dependency. Hou et al. \shortcite{hou} first learn the abstract label transition probabilities in source domains and then copy them to target domains. As Figure \ref{figure2}a shows, the abstract \texttt{O$\rightarrow$I} probability is copied to three targets directly (the red lines). However, this makes the target probability sum of \texttt{O$\rightarrow$}(all labels) end up with 160\%. To avoid the possible probability overflows, Yang and Katiyar \shortcite{yang} propose an even distribution method.
As Figure \ref{figure2}b shows, the abstract \texttt{O$\rightarrow$I} probability is distributed evenly among the three targets (the green lines). However, this could lead to severe contradictions between the target probabilities and reality. For example, there are 4,983 \texttt{DATE} entities and only one \texttt{EMAIL} entity in the I2B2 test set, so the target probabilities of \texttt{O$\rightarrow$I}-\texttt{DATE} and \texttt{O$\rightarrow$I}-\texttt{EMAIL} should be clearly different.}
{Consequently, the current dependency transferring may lead to misclassifications due to the roughly estimated target transition probabilities, even though it sheds some light on few-shot NER.

In addition, the majority of prototypical models for few-shot NER \cite{jiawei,li} obtain prototypes by averaging the embeddings of each class's support examples, while Yoon et al. \shortcite{tapnet} and Hou et al. \shortcite{hou} demonstrate that such prototypes distribute closely in the embedding space, thus often causing misclassifications. }

In this paper, we {aim to} tackle the above issues inherent in token-level prototypical models.
To this end, we propose \textbf{EP-Net}, an \textbf{E}ntity-level \textbf{P}rototypical \textbf{Net}work enhanced by dispersedly distributed prototypes, as Figure \ref{figure1} shows.
EP-Net builds entity-level prototypes and considers text spans as candidate entities. 
Thus it can determine whether a span is an entity directly according to the most similar prototype to the span. This also eliminates the need for the label dependency. 
For example, EP-Net determines the \textit{``rain’’} and \textit{``tonight''} are two entities, and their types are the \texttt{Weather} and \texttt{Time}, respectively (Figure \ref{figure1}).\footnote{We also add a \texttt{None} type and assign it to spans that are not entities.}
In addition, to distribute these prototypes dispersedly, EP-Net trains them using a distance-based loss from scratch.
And EP-Net aligns spans and prototypes in the same embedding space by utilizing a deep neural network to map span representations to the embedding spaces of prototypes.

In essence, EP-Net is a span-based model. 
Several span-based models \cite{lispan,spanner,sner} have been proposed for the supervised NER task. 
Our EP-Net differs from these models in two ways: (1) The EP-Net obtains entities based on the span-prototype similarity, while these models do so by classifying span representations. (2) The EP-Net works effectively with few labeled examples, whereas these models need a large number of labeled examples to guarantee good performance.

We evaluate our EP-Net on the tag set extension and domain transfer tasks, as well as the Few-NERD settings. Experimental results demonstrate that EP-Net consistently achieves new state-of-the-art overall performance. Qualitative analyses ($\S$\ref{5.5}-\ref{5.6}) and ablation studies ($\S$\ref{5.7}) further validate the effectiveness of EP-Net.

In summary, we conclude the contributions as follows: 
(1) As far as we know, we are among the first to propose an entity-level prototypical network for few-shot NER. 
(2) {We propose a prototype training strategy to augment the prototypical network with dispersedly distributed prototypes.}
(3) Our model achieves the current best overall performance on two evaluation tasks and the Few-NERD.

\section{Related Work}
\textbf{Meta Learning.} Meta learning aims to learn a general model that enables us to adapt to new tasks rapidly based on a few labeled examples \cite{li}. 
One of the most typical metric learning methods is the prototypical network \cite{prototypical}, which learns a prototype for each class and classifies an item based on item-prototype similarities. 
Metric learning has been widely investigated for NLP tasks, such as text classification \cite{sunsheng,geng,baoyu}, relation classification \cite{lv,gaotianyu} and NER \cite{jiawei}. However, these methods use the prototypes obtained by averaging the embeddings of support examples for each class, which are closely distributed. 
In contrast, our model uses dispersedly distributed prototypes obtained by supervised prototype training.

\noindent \textbf{Few-shot NER.}
Previous few-shot NER models  \cite{li,tong} generally formulate the task as a sequence labeling task and propose to use the token-level prototypical network. Thus these models call for label dependency to guarantee good performance. However, it is hard to obtain exact dependency since the label sets vary greatly across domains. 
As an alternative, Hou et al. \shortcite{hou} propose to transfer estimated dependency. They copy the learned abstract dependency from source to target domains, but the target dependency contradicts the probability definition.
Yang and Katiyar \shortcite{yang} propose \textit{StructShot}, which improves the above dependency transferring by equally distributing the abstract dependency to target domains, whereas the target dependency contradicts the reality. 
Das et al. \shortcite{das} introduce Contrastive Learning to the \textit{StructShot}, which inherits the estimated dependency transferring.
We demonstrate that the roughly estimated dependency may harm model performance.
In addition, prompt-based models \cite{cuile,sunyi,ppt,ganqu,open} have been widely researched for this task recently, but the model performance heavily relies on the chosen prompts. Current with our work, Wang et al. \shortcite{wangpeiyi} also propose a span-level prototypical network to bypass label dependency, but their work is still hampered by closely distributed prototypes.
In contrast, our model constructs dispersedly distributed entity-level prototypes, thus avoiding the roughly estimated label dependency and closely distributed prototypes.

\section{Task Formulation and Setup}
In this section, we formally define the task and then introduce the standard evaluation setup.

\subsection{Few-shot NER}
We define an unstructured sentence as a token sequence $\mathcal{X}$ $=$ $(x_1, x_2,…, x_n)$, and define entities annotated in $\mathcal{X}$ 
as ${\mathcal{E}} = [(e^{(1)}, t^{(1)}), ..., (e^{(k)}, t^{(k)})]$, where $e^{(i)}$ and $t^{(i)}$ denote entity text and entity type, respectively. A domain $\mathcal{D}=\{(\mathcal{X}^{(i)}, \mathcal{E}^{(i)})\}_{i=1}^{N_\mathcal{D}}$ is a set of ($\mathcal{X}, \mathcal{E}$) pairs, and each $\mathcal{D}$ has a domain-specific entity type set $\mathcal{T}_\mathcal{D}=\{{t}_i\}^N_{i=1}$, and $N$ is various across domains.

We achieve the few-shot task through three steps: \textbf{Train}, \textbf{Adapt} and \textbf{Recognize}. 
We first \textbf{train} EP-Net with the data of source domains $\{\mathcal{D}_1, \mathcal{D}_2, . . .\}$. Then we then \textbf{adapt} the trained EP-Net to target domains \{$\mathcal{D}’_1, \mathcal{D}’_2,…$\} by fine-tuning it on support sets sampled from target domains. Finally, we \textbf{recognize} entities of query sets using the domain-adapted EP-Net.
We formulate a support set as $\mathcal{S}$$=$$\{(\mathcal{X}^{(i)}, \mathcal{E}^{(i)})\}_{i=1}^{N_\mathcal{S}}$, where $\mathcal{S}$ usually includes a few labeled examples ($K$-{shot}) of each entity type.
For a target domain, we formally define the $K${-shot} NER as follows: given a sentence $\mathcal{X}$ and a $K$-shot support set, find the best entity set ${\mathcal{E}}$ for $\mathcal{X}$.

\subsection{The Standard Evaluation Setup} \label{3.2}
To facilitate meaningful comparisons of results for future research, Yang and Katiyar \shortcite{yang} propose a standard evaluation setup. 
The setup consists of the query set and support set constructions. 

\subsubsection{Query Set Construction}\label{3.2.1}
They argue that traditional construction methods sample different entity classes equally without considering entity distributions. For example, the I2B2 test set contains 4,983 \texttt{DATE} entities, while it only contains one \texttt{EMAIL} entity. Thus they propose to use the original test sets of standard NER datasets as the query sets, thus improving the reproducibility of future studies.

\subsubsection{Support Set Construction}\label{3.2.2}
To construct support sets, they propose a Greedy Sampling algorithm to sample sentences from the dev sets of standard NER datasets. In particular, the algorithm samples sentences for entity classes in increasing order of their frequencies. We present the algorithm in {Appendix \ref{A}}.


\begin{figure*}[t]
\centering
\includegraphics[width=0.77\textwidth]{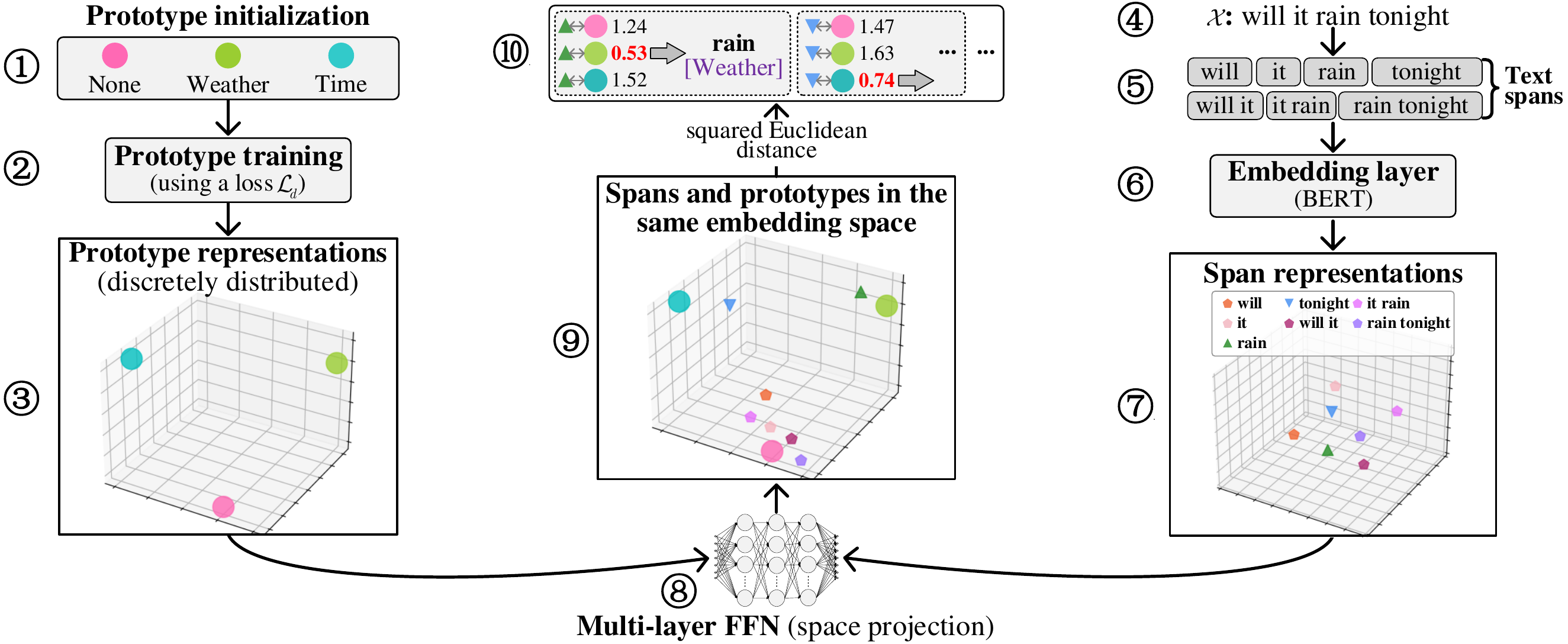} 
\caption{The architecture of EP-Net.  As an example, we use the sentence in Figure \ref{figure1} and 3-dimensional embedding spaces. EP-Net first initializes entity-level prototypes for entity types (\ding{172}). Then it trains the prototypes with a distance-based loss $\mathcal{L}_d$ (\ding{173}), {distributing them dispersedly in embedding space} (\ding{174}). Next, given a sentence $\mathcal{X}$ (\ding{175}), EP-Net first obtains text spans (\ding{176}) and then uses BERT (\ding{177}) to generate span representations (\ding{178}). Fourth, EP-Net uses the space projdection (\ding{179}) to align spans and prototypes in the same embedding space (\ding{180}).
Finally, EP-Net calculates span-prototype similarities measured by the squared {Euclidean distance}. The shorter the distance, the better the similarity. EP-Net classifies entities based on the best similarity, e.g., classifying the ``\textit{rain}'' into the \texttt{Weather} type (\ding{181}).} 
\label{figure3}
\end{figure*}

\section{Model}
In this section, we first provide an overview of EP-Net in $\S$\ref{4.1}, and then illustrate the model initializations in $\S$\ref{4.2} and discuss the model in $\S$\ref{4.3}.

\subsection{EP-Net}\label{4.1}

Figure \ref{figure3} shows the overall architecture of EP-Net.
Given a domain $\mathcal{D}$$=$$\{(\mathcal{X}^{(i)}, \mathcal{E}^{(i)})\}_{i=1}^{N_\mathcal{D}}$ and its entity type set $\mathcal{T}_\mathcal{D}=\{t_i\}_{i=1}^N$, we first initialize an entity-level prototype for each $t_i$ (Figure \ref{figure3}-\ding{172}).
\begin{equation}
\pmb{\it{\Phi}} = \{\pmb{\phi}_0, \pmb{\phi}_1, \pmb{\phi}_2,..., \pmb{\phi}_N\},
\end{equation}
where $\pmb{\it{\Phi}}\in\mathbb{R}^{(N{+}1)*{d_1}}$, and $d_1$ is the dimension of prototype representation. $\pmb{\phi}_0$ is {the} prototype of the \texttt{None} type, and $\pmb{\phi}_i \ (i >0)$ is the prototype of $t_i$.

We design a distance-based loss $\mathcal{L}_d$ to supervise the prototype training (Figure \ref{figure3}-\ding{173}), aiming to distribute these prototypes dispersedly in the embedding space (Figure \ref{figure3}-\ding{174}). 
We argue that the prototypes should be dispersedly distributed in an appropriate-sized embedding space, neither too large nor too small ($\S$\ref{5.5}). Thus we first set a threshold $\tau$ to limit the averaged prototype distance. Then we calculate the squared Euclidean distance between any two prototypes and obtain the averaged prototype distance (denoted as $Euc(\pmb{\it{\Phi}})$). Next, we construct the $\mathcal{L}_d$ as follows.
\begin{subequations}
\begin{normalsize}
\begin{align}
\resizebox{.15\hsize}{!}{$Euc(\pmb{\it{\Phi}})$} &= \resizebox{.65\hsize}{!}{$
\frac{\sum_{i=0}^{N}\sum_{j=0}^{N}\sum_{k=1}^{d} {(\pmb{\it{\phi}}_{i,k}-\pmb{\it{\phi}}_{j,k}})^2}{(N+1)^2}$\small{,}} \\
\psi&=\left\{ \begin{array}{rcl}
\resizebox{.22\hsize}{!}{$Euc(\pmb{\it{\Phi}})-\tau$} & {\textrm{if}}\ \ \  \resizebox{.22\hsize}{!}{$Euc(\pmb{\it{\Phi}})\geq \tau$,} \\
\resizebox{.22\hsize}{!}{$\tau-Euc(\pmb{\it{\Phi}})$}  & {\textrm{if}}\ \ \ \resizebox{.22\hsize}{!}{$Euc(\pmb{\it{\Phi}})<\tau$,} \end{array}\right.\\
\mathcal{L}_d &= \log(\psi+1) \label{2c}.
\end{align}
\end{normalsize}
\end{subequations}
\indent The training goal is to achieve $\psi\rightarrow0^+$, which equals to ${(\psi+1)}\rightarrow1^+$. Thus we design the $\log(\cdot)$ loss (Eq.\ref{2c}), where the smaller the $\mathcal{L}_d$, the more the $(\psi+1)\rightarrow1^+$.

Next, given a sentence $\mathcal{X}=\{x_i\}_{i=1}^{n}$ of the domain $\mathcal{D}$, we first obtain text spans (denoted as $s$, Figure \ref{figure3}-\ding{175}\ding{176}):
\begin{equation}\label{eq3}
s =\{x_i, x_{i+1},..., x_{i+j}\} \ \resizebox{.4\hsize}{!}{$s.t. \ 1 \leq i \leq i+j \leq n$,}
\end{equation}

\noindent where the span length $j$+$1$ is limited by a threshold $\epsilon$: $j$+$1\leq\epsilon$.
To obtain the span representation (Figure \ref{figure3}-\ding{177}\ding{178}), we first use BERT \cite{bert} to generate the embedding sequence of $\mathcal{X}$. 
\begin{equation}
\mathbi{H}_\mathcal{X} = \{\pmb{h}_1, \pmb{h}_2,…, \pmb{h}_n\},
\end{equation}
 
\noindent where $\mathbi{H}_\mathcal{X}\in \mathbb{R}^{n*d_2}$, and $d_2$ is the BERT embedding dimension. $\pmb{h}_i$ is the BERT embedding of $x_i$.
We use $\mathbi{H}_s$ to denote the BERT embedding sequence of span $s$.
\begin{equation}
\mathbi{H}_{s} = \{\pmb{h}_i, \pmb{h}_{i+1},…, \pmb{h}_{i+j}\}.
\end{equation}

We obtain the span representation (denoted as $\mathbi{E}_s$) by concatenating the max-pooling of $\mathbi{H}_s$ (denoted as $\tilde{\mathbi{H}}_s$) and the span length embedding.
\begin{subequations}
\begin{normalsize}
\begin{align}
\tilde{\mathbi{H}}_s &=\resizebox{.85\hsize}{!}{$ [\max(\pmb{h}_{i,1},..., \pmb{h}_{i+j,1}),..., \max(\pmb{h}_{i,d_1},..., \pmb{h}_{i+j,d_1})]$,} \\
\mathbi{E}_{s} &= [\tilde{\mathbi{H}}_s; \pmb{w}_{j+1}]\label{6b},
\end{align}
\end{normalsize}
\end{subequations}
{where} $\tilde{\mathbi{H}}_s\in\mathbb{R}^{d_2}$, $\mathbi{E}_{s} \in \mathbb{R}^{d_2+d_3}$. $\pmb{w}_{j+1}$ is the length embedding trained for spans with a length $j$+$1$ and $d_3$ is the embedding dimension.
Due to the fact that $\mathbi{E}_{s}$ and prototype representations ($\pmb{\it{\Phi}}$) are not in the same embedding space, we project $\mathbi{E}_{s}$ to the embedding space of $\pmb{\it{\Phi}}$ using a multi-layer Feed Forward Network (FFN)\footnote{The FFN enables us to fine-tune our model on support sets without overfitting due to its simple neural architecture.}
 and denote the aligned span representation as $\tilde{\mathbi{E}}_s$ (Figure \ref{figure3}-\ding{179}\ding{180}).
\begin{equation}\label{eq7}
\tilde{\mathbi{E}}_{s} = \mathbi{E}_{s} \mathbi{W} + \mathbi{b},
\end{equation}
where $\tilde{\mathbi{E}}_{s} \in \mathbb{R}^{d_1}$, $\mathbi{W}$ and $\mathbi{b}$ are FFN parameters.
Next, for the span $s$, we calculate the similarity between it and each prototype ${\pmb{\phi}_i}\in \pmb{\it\Phi}$ using the squared Euclidean distance.
\begin{equation}\label{eq8}
\textrm{sim}(s, {\pmb\phi_i})_{i=0}^N = \sum_{j=1}^d(\tilde{\mathbi{E}}_{s,j}-\pmb\phi_{i,j})^2.
\end{equation}

As a shorter distance denotes a better similarity, we classify entities based the shortest distance (Figure \ref{figure3}-\ding{181}):
\begin{equation}\label{eq9}
{t}_s = \mathrm{arg}\ \underset{\pmb{\phi}_i \in \pmb{\it\Phi}}{\mathrm{min}} \ \textrm{sim}(s, {\pmb{\it\phi_i}})_{i=0}^N,
\end{equation}
where ${t}_s\in\mathcal{T}_\mathcal{D}$ is the type classified for the span $s$.

To construct an entity classification loss, we first take the $-\textrm{sim}(s, {\pmb\phi_i})_{i=0}^N$ as classification logits, thus the best similarity has the largest logit. We then normalize these logits using the {softmax} function. Finally, we construct a cross-entropy loss $\mathcal{L}_s$.
\begin{subequations}\label{eq10}
\begin{normalsize}
\begin{align}
\pmb{\hat{y}}_{s,i} &= \frac{\exp^{-\textrm{sim}(s, \pmb\phi_i)}}{\sum_{j=0}^{N}{\exp^{-\textrm{sim}(s, \pmb\phi_j)}}}, \\
\mathcal{L}_s &= -\frac{1}{M_{s}} \sum_{j=1}^{M_{s}}\sum_{i=0}^{N}{\pmb y}^j_{{s,i}}\log\hat{\pmb y}^j_{{s,i}}\label{eq10b},
\end{align}
\end{normalsize}
\end{subequations}
where $\{\pmb{{y}}_{{s}}, \pmb{{\hat{y}}}_{{s}}\}\in\mathbb{R}^{N+1}$, and $\pmb{{y}}_{{s}}$ is the one-hot vector of the gold type for the span ${{s}}$. ${M_{s}}$ is the number of span instances.

During model training, we optimize model parameters by minimizing the following joint loss.
\begin{equation}
\mathcal{L}(\it{W};{\theta}) = \mathcal{L}_d + \mathcal{L}_s.
\end{equation}

\subsection{Initializations}\label{4.2}
\subsubsection{Train Initialization}
In the \textbf{Train} step, given a source domain $\mathcal{D}$ with an entity type set $\mathcal{T}_\mathcal{D}$$=$$\{{t}_i\}^N_{i=1}$, we randomly initialize the entity-level prototypes $\pmb{\it{\Phi}}=\{\pmb{\phi}_i\}_{i=0}^N$.We assign $\pmb{\phi}_0$ to the \texttt{None} type and $\pmb{\phi}_i$ to $t_i$. To guarantee that we can adapt EP-Net to target domains that have more types than the domain $\mathcal{D}$, we actually initialize $\pmb{\it{\Phi}}=\{\pmb{\phi}_i\}_{i=0}^{100}$, where EP-Net can be adapted to any target domains with entity types less than 100. Moreover, the $N$ can be set to an even larger value if necessary. By doing so, the prototypes $\{\pmb{\phi}_i\}_{i=N+1}^{100}$ are unassigned, but we can still distribute them dispersedly through training them using the loss $\mathcal{L}_d$ ($\S$\ref{5.6}).
We use the bert-base-cased model in the embedding Layer.\footnote{\url{{https://huggingface.co/bert-base-uncased.}}}

\subsubsection{Adapt and Recognize Initializations}
In the \textbf{Adapt} step, given a target domain $\mathcal{D'}$ with an entity type set $\mathcal{T}_{\mathcal{D}'}$$=$$\{{{t}'_i}\}^{N'}_{i=1}$ and the EP-Net trained in the \textbf{Train} step, we first assign a prototype of the trained $\pmb{\it{\Phi}}=\{\pmb{\phi}_i\}_{i=0}^{100}$ to each $t'_i$. In particular, we assign $\pmb{\phi}_0$ to the \texttt{None} type. And if there are types that are overlapped between $\mathcal{T}_\mathcal{D}$ and $\mathcal{T}_\mathcal{D'}$ (i.e., $\mathcal{T}_\mathcal{D} \cap \mathcal{T}_\mathcal{D'}\neq \varnothing$), for each overlapped type, we reuse the prototype assigned in the \textbf{Train} step. For other types in $\mathcal{T}_\mathcal{D'}$, we randomly assign an unassigned prototype in $\pmb{\it{\Phi}}$ to it, and we first choose the prototype that is ever assigned in the \textbf{Train} step.
Then, we adapt EP-Net to the domain $\mathcal{D}'$ by fine-tuning it on support sets sampled from $\mathcal{D'}$.

However, {Fine-tuning} the model with small support sets runs the risk of overfitting. To avoid this, we propose to use the following strategies: (1) We freeze the BERT and solely fine-tune the assigned prototypes and the multi-layer FFN. (2) We use an early stopping criterion, where we continue fine-tuning our model until the loss starts to increase. (3) We set upper limits for fine-tuning steps, where the model will stop when reaching the limits even though the loss continues decreasing.
With the above strategies, we demonstrate that only a few fine-tuning steps on these examples can make rapid progress without overfitting.

In the \textbf{Recognize} step, we use the domain-adapted EP-Net to recognize entities in the query set of $\mathcal{D'}$ directly.

\subsection{Model Discussion}\label{4.3}
In the \textbf{Train} step, the randomly initialized prototypes cannot represent entity types at first. Through the joint model training with the $\mathcal{L}(\it{W};{\theta})$, EP-Net establishes correlations between entity types and their assigned prototypes. Moreover, the multi-layer FFN can also be trained to cluster similar spans around related prototypes in the embedding space. As Figure \ref{figure3}-\ding{180} shows, the ``\textit{rain}'' is mapped to be closer to the \texttt{Weather} than other prototypes.

To precisely simulate the few-shot scenario, we are not permitted to count the entity length of target domains. Thus we set the span length threshold $\epsilon$ to an empirical value of 10 based on source domains. For example, 99.89\% of the entities in the OntoNotes have lengths under 10. 


We propose a heuristic method for removing overlapped entities classified by EP-Net. Specifically, we keep the one with the best span-prototype similarity of those overlapped entities and drop the others.

Concurrently, Wang et al. \shortcite{wangpeiyi} propose a  span-level model -- ESD. We summarize how our EP-Net differs from the ESD as follows:
(1) Our EP-Net fine-tunes on support sets while the ESD solely uses them for similarity calculation without fine-tuning. Ma et al. \shortcite{matingting} claim that the fine-tuning method is far more effective in using the limited information in support sets.
(2) The ESD obtains class prototypes with embeddings of the same classes in support sets, thus suffering from closely distributed prototypes \cite{hou}.  By contrast, our EP-Net avoids this by training dispersedly distributed prototypes from scratch.

\begin{table*}[h] \small
\centering
\renewcommand\tabcolsep{2pt}
\begin{tabular}{llllllllllllll}
\toprule
\multirow{2}{*}{\textbf{Model}} & \multicolumn{4}{c}{\textbf{Tag Set Extension}}                   & \textbf{} & \multicolumn{4}{c}{\textbf{Domain Transfer}}    & \textbf{} & \multicolumn{3}{c}{\textbf{Few-NERD}}\\ \cmidrule{2-5} \cmidrule{7-10}  \cmidrule{12-14}
 & Group A           & Group B           & Group C           & Avg. &           & I2B2      & CoNLL & WNUT              & Avg.    && Intra & Inter              & Avg. \\ \midrule
ProtoNet    		& 18.7$\pm$4.7 		& 24.4$\pm$8.9  		& 18.3$\pm$6.9  & 20.5 &  &\ \ 7.6$\pm$3.5  &53.0$\pm$7.2    & 14.8$\pm$4.9    &25.1    &&18.6$\pm$7.2 &25.3$\pm$8.8 & \\
ProtoNet+P\&D \   & 18.5$\pm$4.4 & 24.8$\pm$9.3  & 20.7$\pm$8.4  & 21.3 &  &\ \ 7.9$\pm$3.2  & 56.0$\pm$7.3   & 18.8$\pm$5.3    &27.6    &&19.4$\pm$5.6 &26.2$\pm$4.2 &22.8   \\
NNShot                 & 27.2$\pm$3.5 & 32.5$\pm$14.4 & 23.8$\pm$10.2 & 25.7 &  & 16.6$\pm$2.1  & 61.3$\pm$11.5   & 21.7$\pm$6.3    &33.2    &&20.1$\pm$8.5 &25.7$\pm$7.7 &22.9  \\
StructShot             &27.5$\pm$4.1 & 32.4$\pm$14.7 & 23.8$\pm$10.2 & 27.9 &  & 22.1$\pm$3.0  & 62.3$\pm$11.4   & 25.3$\pm$5.3    & 36.6   &&20.3$\pm$4.3 &26.7$\pm$5.6 &23.5  \\ 
CONTaiNER             & 32.4$\pm$5.1 & 30.9$\pm$11.6 & 33.0$\pm$12.8 & 32.1 &  &  21.5$\pm$1.7  &  61.2$\pm$10.7  & 27.5$\pm$1.9    &36.7    &&22.4$\pm$5.4 &28.4$\pm$4.3 &25.4  \\ 
\midrule
EP-Net (ours)         & \textbf{38.4}$\pm$4.5 & \textbf{42.3}$\pm$10.8  & \textbf{36.7}$\pm$9.5  &\textbf{39.1}      &  & \textbf{27.5}$\pm$4.6  & \textbf{64.8}$\pm$10.4  & \textbf{32.3}$\pm$4.8    &\textbf{41.5}       &&\textbf{25.8}$\pm$5.1 &\textbf{30.9}$\pm$4.9 &\textbf{28.4}   \\ \bottomrule
\end{tabular}
\caption{F1 scores of 1-shot experiments. We report the mean and standard deviations of F1 scores. }
\label{table1}
\end{table*}

\begin{table*}[] \small
\centering
\renewcommand\tabcolsep{2pt}
\begin{tabular}{llllllllllllll}
\toprule
\multirow{2}{*}{\textbf{Model}} & \multicolumn{4}{c}{\textbf{Tag Set Extension}}                   & \textbf{} & \multicolumn{4}{c}{\textbf{Domain Transfer}}    & \textbf{} & \multicolumn{3}{c}{\textbf{Few-NERD}}\\ \cmidrule{2-5} \cmidrule{7-10}  \cmidrule{12-14}
                                & Group A           & Group B           & Group C           & Avg. &           & I2B2      & CoNLL & WNUT              & Avg.    && Intra & Inter              & Avg. \\ \midrule
ProtoNet            & 27.1$\pm$2.4 \ \          & 38.0$\pm$5.9   \  \     & 38.4$\pm$3.3   \   \     & 34.5 &           & 10.3$\pm$0.4          & 65.9$\pm$1.6  \  \ & 19.8$\pm$5.0     &32.0  & &33.2$\pm$6.4   &31.7$\pm$5.9  &  \\
ProtoNet+P\&D            & 29.8$\pm$2.8          & 41.0$\pm$6.5          & 38.5$\pm$3.3          & 36.4 &           & 10.1$\pm$0.9     & 67.1$\pm$1.6     & 23.8$\pm$3.9          & 33.6 & & 26.4$\pm$3.8   &28.7$\pm$7.2  &27.6 \\
NNShot                          & 44.7$\pm$2.3          & 53.9$\pm$7.8          & 53.0$\pm$2.3          & 50.5 &           & 23.7$\pm$1.3          & 74.3$\pm$2.4    & 23.9$\pm$5.0     &40.7 & & 29.6$\pm$5.3 &33.9$\pm$5.1 &31.8 \\
StructShot                      & 47.4$\pm$3.2          & 57.1$\pm$8.6            & 54.2$\pm$2.5                  & 52.9 &           & 31.8$\pm$1.8          & 75.2$\pm$2.3    & 27.2$\pm$6.7     &44.7  & & 31.2$\pm$4.4 &35.7$\pm$3.8  &33.5  \\ 
CONTaiNER                     & 51.2$\pm$6.0         & 56.0$\pm$6.2          & \textbf{61.2}$\pm$2.7          & {56.2} &           & 36.7$\pm$2.1       & 75.8$\pm$2.7          & 32.5$\pm$3.8          &48.3  & & 33.1$\pm$4.6 &$38.4\pm$4.4  &35.8\\

\midrule
EP-Net (ours)                  & \textbf{55.5}$\pm$3.2 & \textbf{64.8}$\pm$4.8 & {52.7}$\pm$2.2 & \textbf{57.7}     &           & \textbf{44.9}$\pm$2.7 & \textbf{78.8}$\pm$2.7  & \textbf{38.4}$\pm$5.2 & \textbf{54.0} & & \textbf{36.4}$\pm$4.6   &\textbf{41.4}$\pm$3.6 &\textbf{38.9}     \\ \bottomrule
\end{tabular}
\caption{F1 scores of 5-shot experiments. We report the mean and standard deviations of F1 scores. }
\label{table2}
\end{table*}

\section{Experiments}

\subsection{Evaluation Tasks}
We evaluate EP-Net on two evaluation tasks and the Few-NERD settings using 1- and 5-shot settings. Limited by space, we solely report the key points here and discuss more details in {Appendix \ref{B}}.

\noindent\textbf{Tag Set Extension.} This task aims to evaluate models for recognizing new types of entities in existing domains. 
Yang and Katiyar \shortcite{yang} divide the 18 entity types of the OntoNotes \cite{ontonotes} into three target sets, i.e., {Group A}, {B} and {C}, to simulate this scenario. 
Models are evaluated on one target set while being trained on the others.

\noindent\textbf{Domain Transfer.} This task aims to evaluate models for adapting to a different domain. Yang and Katiyar \shortcite{yang} propose to use the general domain as the source domain and test model on medical, news, and social domains.

\noindent\textbf{Few-NERD Settings.}
Few-NERD \cite{dingning} is a large-scale dataset for few-shot NER. It consists of two different settings: \textbf{Intra} and \textbf{Inter}. The Intra divides the train/dev/test according to coarse-grained types. The Inter divides the train/dev/test according to fine-grained types. Thus the coarse-grained entity types are shared. 
The Intra is more challenging as the restrictions of sharing coarse-grained types.

\subsection{Datasets and Baselines}
For a fair comparison, we use the same datasets and baselines reported in \cite{yang,dingning,das}. Specifically, we use  OntoNotes (general domain),  CoNLL 2003 \cite{conll2003} (news domain), I2B2 2014 \cite{i2b2} (medical domain) and WNUT  2017 \cite{wnut} (social domain) for the tag set extension and domain transfer tasks. 

We compare the performance of EP-Net with previous best models, including: {Prototypical Network} (\textbf{ProtoNet}) \cite{prototypical}, \textbf{ProtoNet+P\&D}  \cite{hou} , \textbf{NNShot} and \textbf{StructShot} \cite{yang} and \textbf{CONTaiNER} \cite{das}.
We represent more baseline details in {Appendix \ref{C}}.

\subsection{Implementation Details}\label{5.3}
In all experiments, we optimize EP-Net using AdamW with a learning rate of 5$e$-5 and set $d_1$ and $d_3$ to 512 and 25, respectively. $d_2$ is 768 when using the BERT base model. We set 3 layers for the multi-layer FFN, the distance threshold $\tau$ to 2, and the train batch size to 2 and 8 in 1- and 5-shot experiments, respectively.
We set the distance threshold $\tau$ to 2 and 3 for 1- and 5-shot experiments, respectively. Moreover, we investigate the model performance against different $\tau$ values in {Appendix} \ref{D}.
Following supervised span-based work \cite{ji}, we sample spans of the \texttt{None} type during model training and set the sampled count to 20 and 40 in 1- and 5-shot experiments, respectively. 
Following \cite{yang,das}, we sample 5 support sets and report the mean and standard deviation of the F1 scores in each experiment.

\subsection{Main Results}

We report experimental results for 1- and 5-shot settings in Table \ref{table1} and Table \ref{table2}, respectively. 
We have the following observations.

(1) In terms of the overall metric (i.e., \textbf{Avg.}), EP-Net consistently outperforms the listed baselines on the two tasks and Few-NERD, delivering +1.5\% to +7.0\% averaged F1 gains. Moreover, EP-Net improves up to +11.4\% F1 scores on 1-shot Group B.
We attribute these gains to the advantages of the proposed entity-level prototypical network.

(2) On the 5-shot Group C, EP-Net is inferior to CONTaiNER by 8.5\% F1 scores. Detailed error analysis indicates that the group's \texttt{DATE} type should bear the primary responsibility. Of the 4,178 entities in the test set, 1,536 are \texttt{DATE} entities, in which there are up to 429 different expressions, such as ``week'', ``this week'', ''last week'', ``2 weeks'', ``2 - week'' etc. However, the 5-shot setting solely enables us to sample very few various expressions, leading to the poor performance in \texttt{DATE} entities. For example, if a support set solely samples the ``week'', it is hard for EP-Net to recognize entities like ``this week'' and ``last week''.

In addition, we conduct episode evaluations on Few-NERD and report the results in {Appendix} \ref{E}.


\subsection{Visualization} \label{5.5}

We use the 1-shot Group A experiment to investigate prototype distributions. Specifically, in the \textbf{Train} step we initialize the prototype set $\pmb{\it{\Phi}}=\{\pmb\phi\}_{i=0}^{100}$ and assign $\{\pmb\phi\}_{i=0}^{12}$ to the \texttt{None} type and the 12 pre-defined entity types of the source domain. In the \textbf{Adapt} step, we assign $\{\pmb\phi\}_{i=0}^{6}$ to the \texttt{None} type and the 6 pre-defined entity types of the Group A. 

\begin{figure}[h]
\centering
{\includegraphics[width=0.489\textwidth]{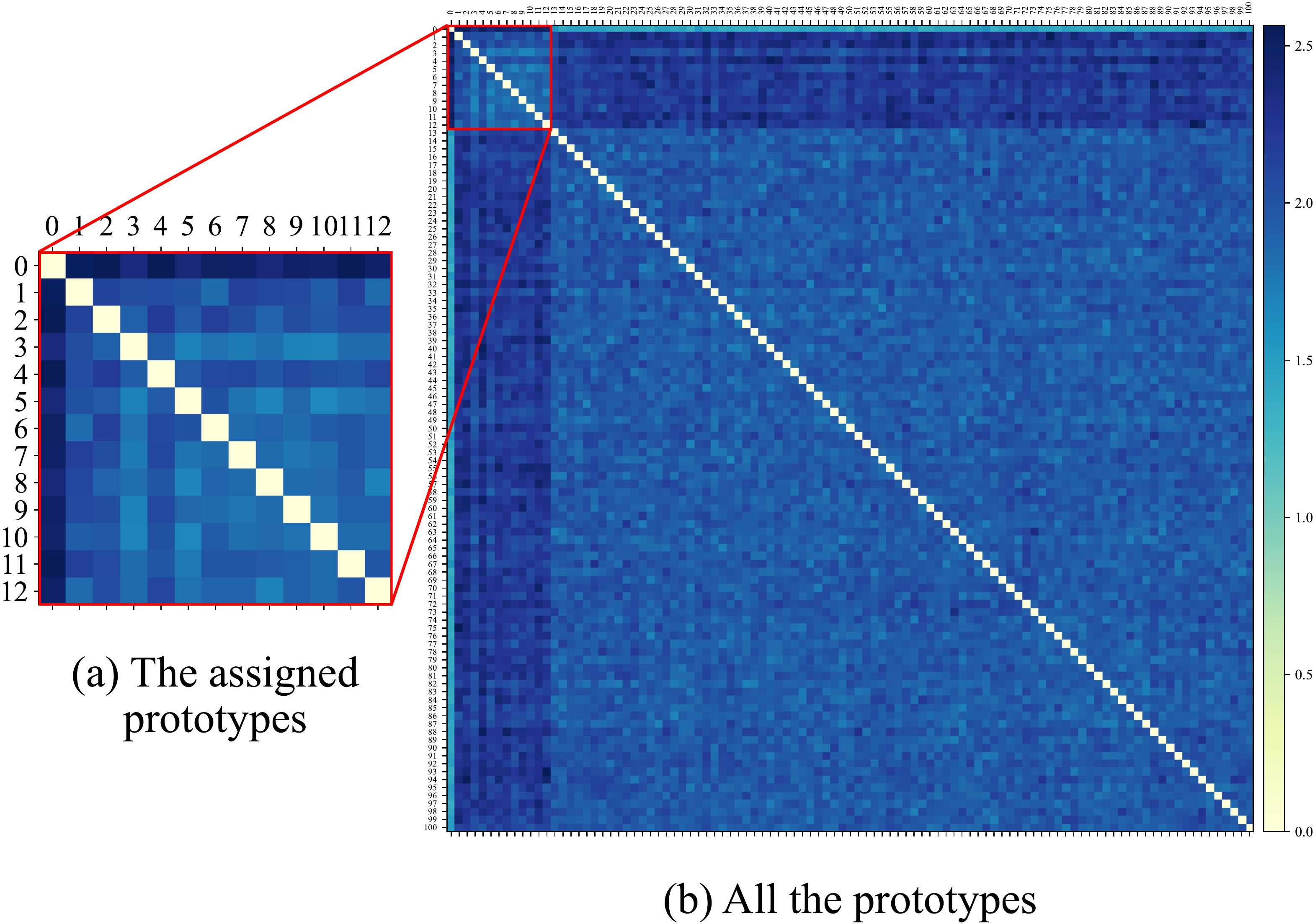} }
\caption{Heat maps of prototype distributions in the embedding space, which are measured by the squared Euclidean distance. In the (b), we show the distributions of all the prototypes $\pmb{\it{\Phi}}=\{\pmb\phi\}_{i=0}^{100}$. In the (a), we amplify the distributions of the 13 assigned prototypes $\{\pmb\phi\}_{i=0}^{12}$.
The darker the color, the larger the distance.}
\label{figure4}
\end{figure}

We report the visualization results in Figure \ref{figure4}.
From Figure \ref{figure4}b, we observe that all prototypes are dispersedly distributed because the Euclidean distance between any two prototypes is approximate 2. Therefore, we conclude that EP-Net can distribute the prototypes dispersedly through the prototype training.
From Figure \ref{figure4}a, we see that the distances between the \texttt{None} type and other assigned prototypes are generally larger than other distances.
We attribute it to the fact that the \texttt{None} type does not represent any unified semantic meaning, thus the \texttt{None} spans actually correspond to a variety of semantic spaces, requiring the \texttt{None}  prototype to keep away from other prototypes to alleviate the misclassification problem.

Moreover, we realize another entity-level prototypical network with conventional prototypes\footnote{We obtain the conventional prototypes by averaging the embeddings of each type’s examples. For the \texttt{None} type, we obtain its prototype by averaging representations of the sampled \texttt{None} spans.\label{note1}}, and refer to it as \textbf{CP-Net}. We do not train the conventional prototypes with the loss $\mathcal{L}_d$ but fine-tune it during the model training. We report more details of CP-Net in {Appendix} \ref{F}.

\begin{figure}[h] 
\centering            
\subfigure[Our prototypes]{
\begin{minipage}[t]{0.20\textwidth}
\centering
\includegraphics[width=1.25in]{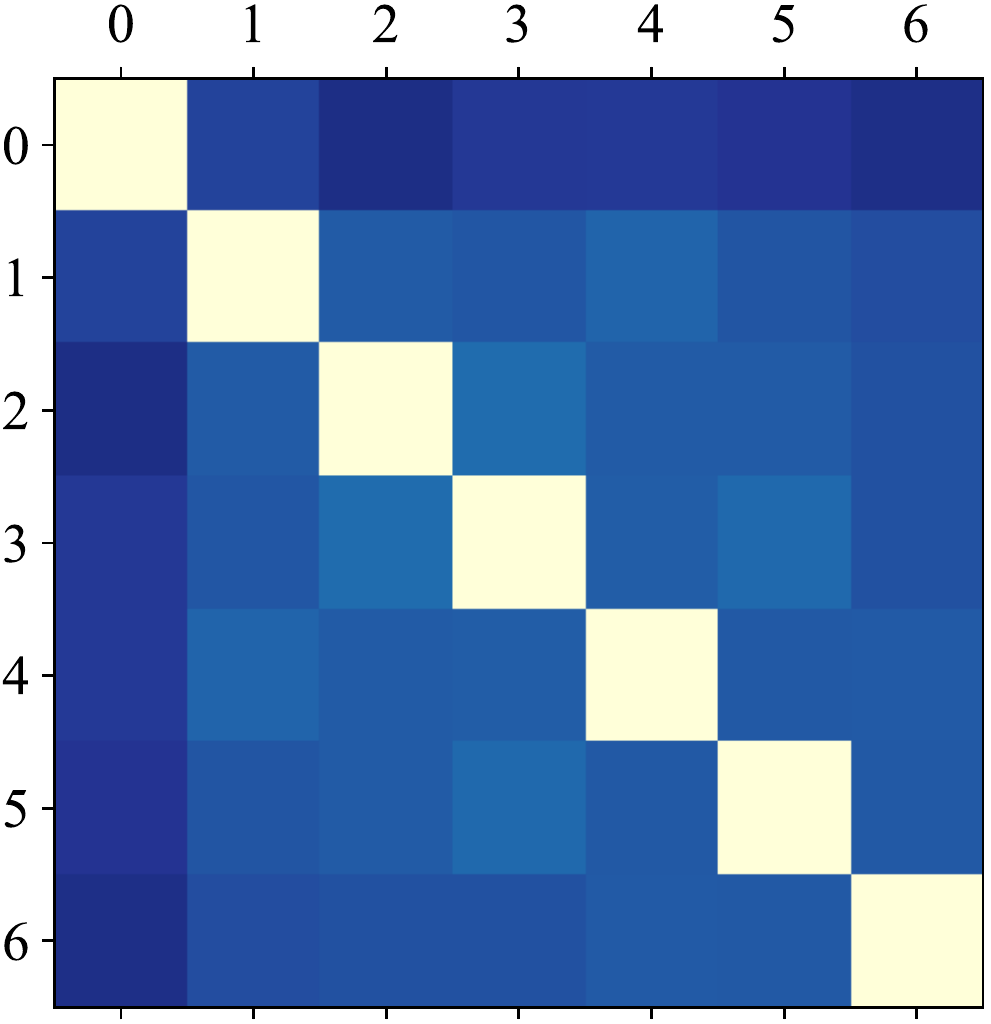}
\end{minipage}
}%
\subfigure[Conventional prototypes]{
\begin{minipage}[t]{0.25\textwidth}
\centering
\includegraphics[width=1.463in]{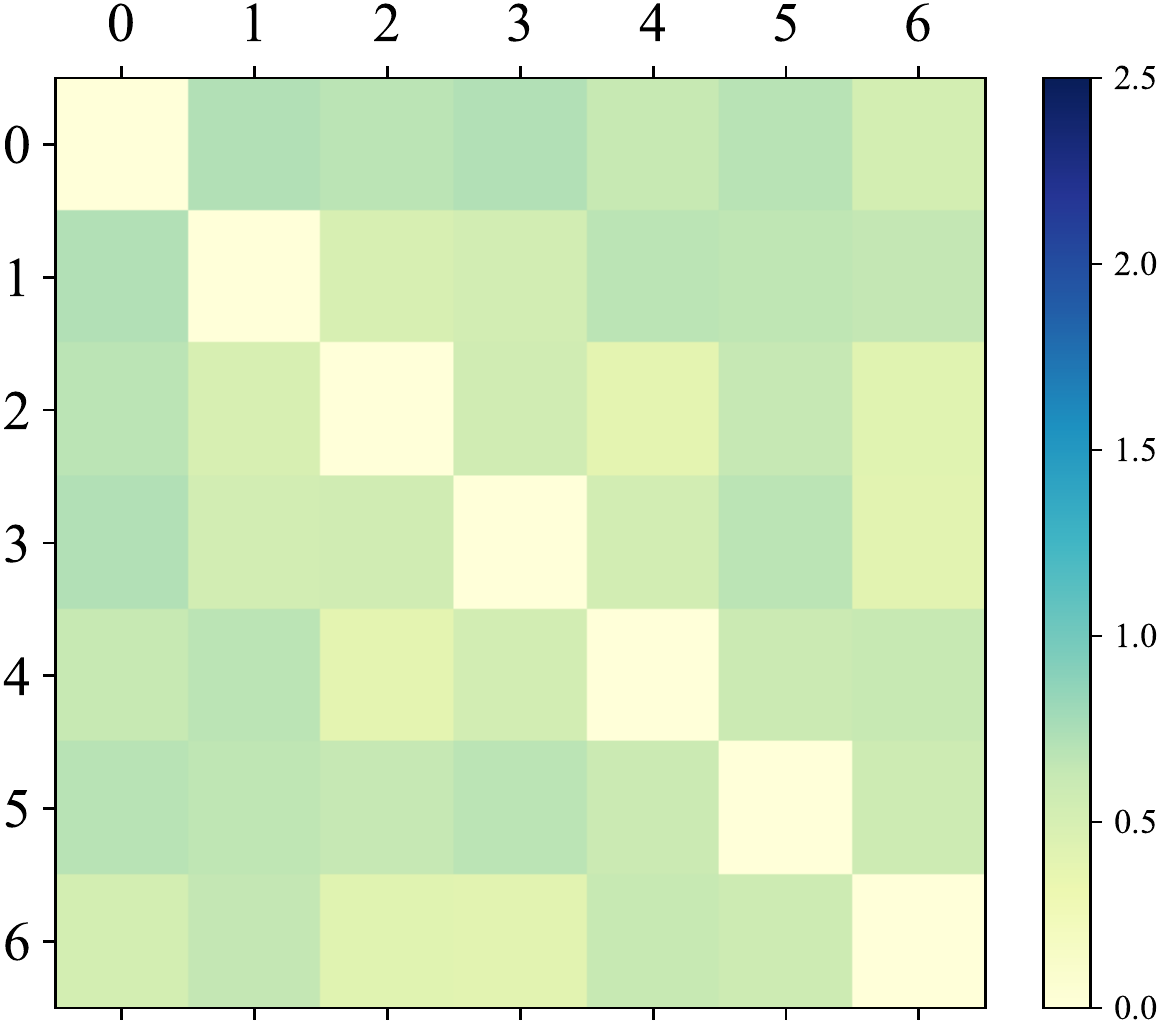}
\end{minipage}
}%
\centering
\caption{Heat maps of prototype distributions of our prototypes and conventional prototypes.}\label{figure5}
\end{figure}

We visualize the distributions of our prototypes and conventional prototypes in Figure \ref{figure5}. 
To be specific, we use prototypes obtained in the \textbf{Recognized} step of both models.
We observe that: (1)  Our prototypes are distributed much more dispersedly than the conventional prototypes. (2) Our \texttt{None} prototype is more distant from other prototypes, whereas the conventional \texttt{None} prototype stays close to other conventional prototypes. 
These results indicate that our prototypes enable us to alleviate the misclassifications caused by closely distributed prototypes.


\subsection{How does the Dispersedly Distributed Prototyes Enhance the EP-Net?} \label{5.6}
We run the 1-shot Group A experiment with EP-Net and CP-Net to conduct the investigation. We first compare the F1 scores of the two models. The results show our EP-Net outperforms CP-Net by +9.6\% F1 scores, verifying the effectiveness of the dispersedly distributed prototypes.

\begin{figure}[] 
\centering
\subfigure[EP-Net]{
\begin{minipage}[t]{0.221\textwidth}
\centering
\includegraphics[width=1.51in]{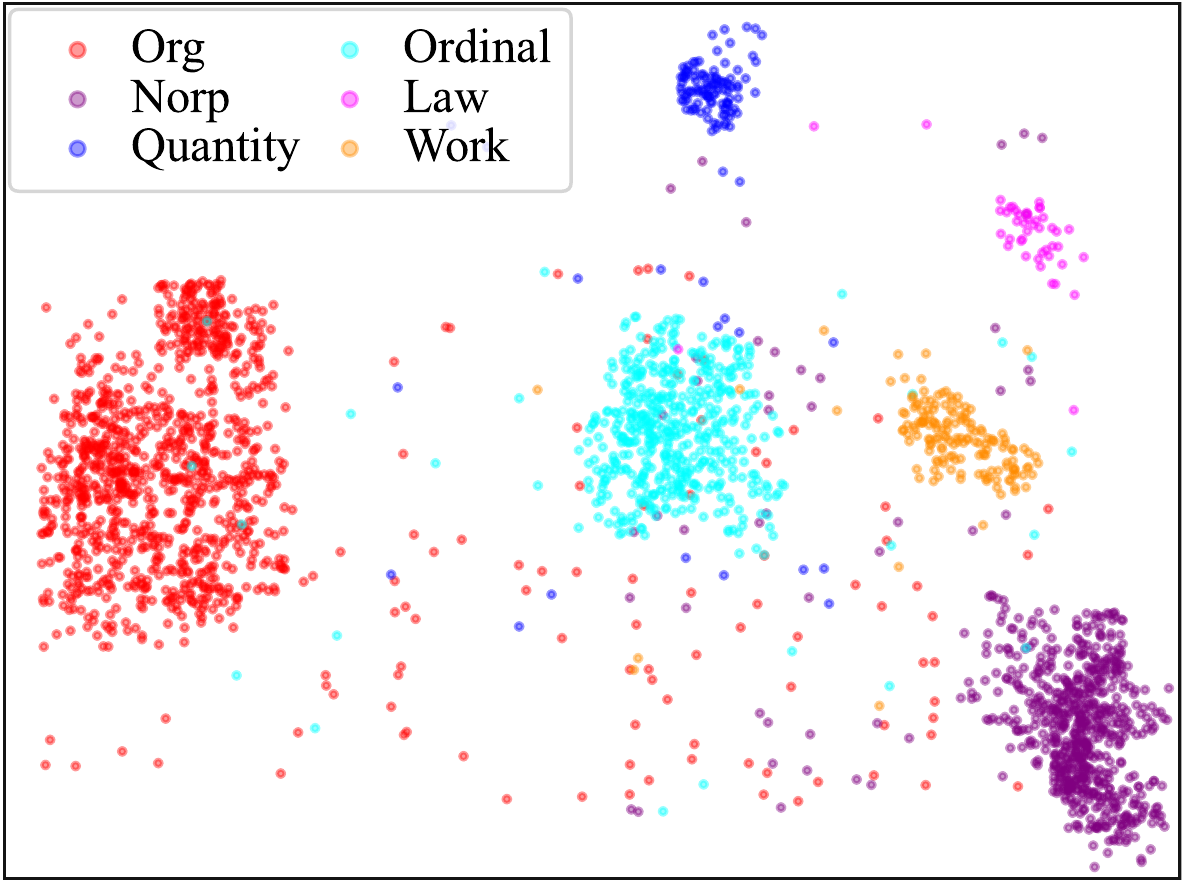}
\end{minipage}%
}%
\subfigure[CP-Net]{
\begin{minipage}[t]{0.27\textwidth}
\centering
\includegraphics[width=1.506in]{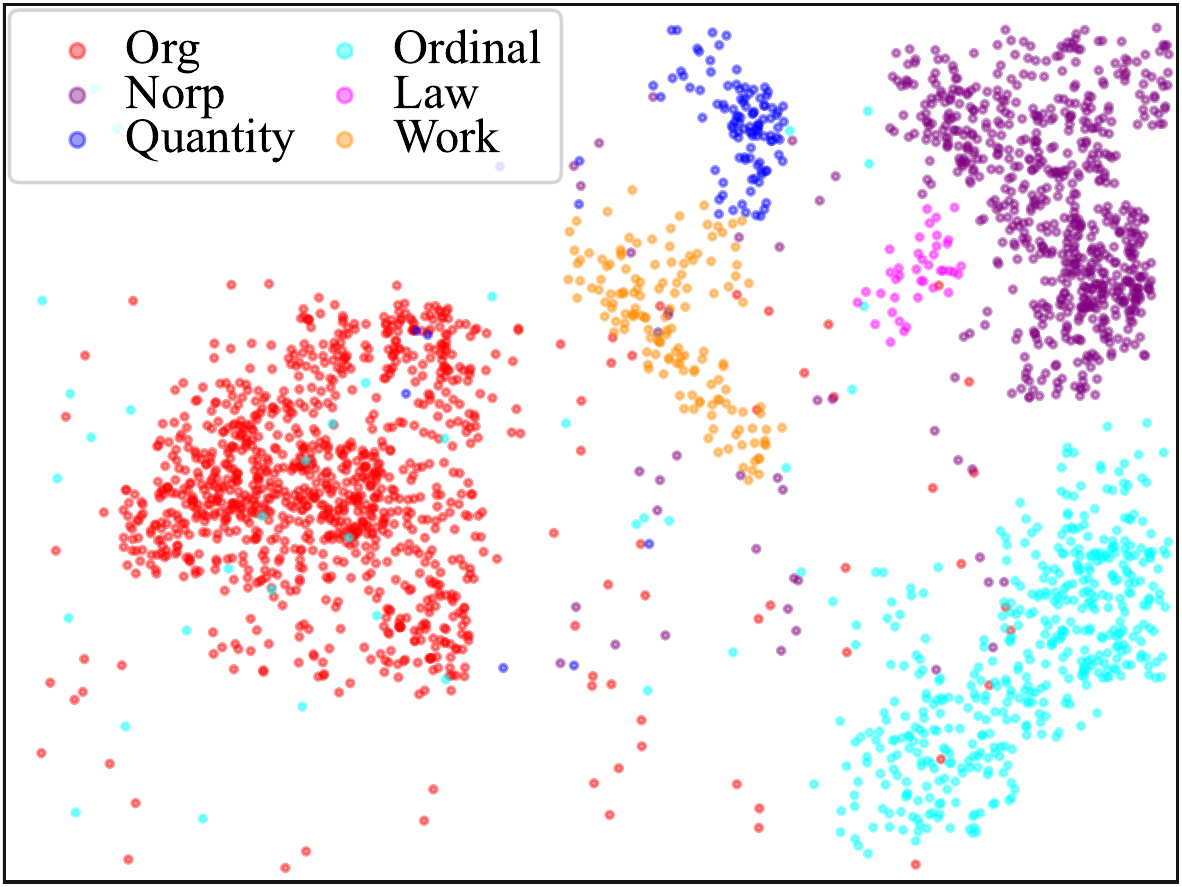}
\end{minipage}%
}%
\centering
\caption{t-SNE visualization of span representations of EP-Net and CP-Net. We obtain these representations in the \textbf{Recognize} step of both models. Since there are too many \texttt{None} spans (890,000+), we do not show their visualizations in the figure.
}\label{figure6}
\end{figure}

In addition, we use t-SNE \cite{tsne} to reduce the dimension of span representations obtained in the \textbf{Recognize} step of EP-Net and CP-Net and visualize these representations in Figure \ref{figure6}. We can see that our EP-Net clusters span representations of the same entity class while dispersing span representations of different entity classes obviously, which we attribute to the usage of dispersedly distributed prototypes. 
Based on the above fact, we conclude that our EP-Net can greatly alleviate the misclassifications caused by closely distributed prototypes. 

\subsection{Ablation Study} \label{5.7}
We conduct ablation studies to investigate the significance of model components 
and report the results in Table \ref{table3}. Specifically,
(1) In the ``{-}{ Entity}$\mbox{-}${level prototype}'', we ablate the entity-level prototypes and use token-level prototypes instead. Moreover, we use the copying method (Figure \ref{figure2}) to transfer the label dependency. The ablation results show that the F1 scores drop from 5.1\% to 7.2\%, validating the advantages of entity-level prototypes.
(2) In the {``$\mbox{-}$ Prototype training''}, we remove the loss $\mathcal{L}_d$ from the $\mathcal{L}(\it{W};{\theta})$, thus the prototypes are not trained being dispersedly distributed. The decreasing F1 scores (5.8\% to 11.8\%) demonstrate that EP-Net significantly benefits from the dispersedly distributed prototypes.
(3) In the ``{$\mbox{-}$Euclidean distance}'', we use the cosine similarity to measure span-prototype similarities instead.
We see that the Euclidean similarity consistently surpasses the cosine similarity, revealing that a proper measure is vital to guarantee good performance, which is consistent with the conclusion in \cite{prototypical}.

\begin{table}[] \small
\centering
\renewcommand\tabcolsep{3.5pt}
\begin{tabular}{lcccc}
\toprule
\textbf{Model} & \multicolumn{1}{c}{\textbf{\begin{tabular}[c]{@{}c@{}}Group A\\ (F1)\end{tabular}}} & \multicolumn{1}{c}{\textbf{\begin{tabular}[c]{@{}c@{}}I2B2\\ (F1)\end{tabular}}} & \multicolumn{1}{c}{\textbf{\begin{tabular}[c]{@{}c@{}}Intra\\ (F1)\end{tabular}}} & \multicolumn{1}{c}{\textbf{\begin{tabular}[c]{@{}c@{}}Inter\\ (F1)\end{tabular}}} \\ \midrule
EP-Net      & 38.4                           & 27.5         & 25.8    & 30.9    \\
- {Entity-level prototype} &    31.6                                 &  22.4            &  18.6     &  25.1          \\
- {Prototype training} &   30.3                                  & 19.8         &  20.0      &  19.1     \\
- {Euclidean distance} &   33.4                                  & 25.2         &  21.6     &  27.3            \\ 
\bottomrule
\end{tabular}
\caption{Ablation results under the 1-shot setting. We select one dataset for each of the two evaluation tasks, as well as the Intra and Inter of the Few-NERD.}\label{table3}
\end{table}

\section{Conclusion}
In this paper, we propose an entity-level prototypical network for few-shot NER (\textbf{EP-Net}). And we augment EP-Net with dispersedly distributed prototypes. The entity-level prototypes enable EP-Net to avoid suffering from the roughly estimated label dependency brought by abstract dependency transferring. Moreover, EP-Net distributes the prototypes dispersedly via supervised prototype training and maps spans to the embedding space of the prototypes to eliminate the alignment biases. Experimental results on two evaluation tasks and the Few-NERD settings demonstrate that EP-Net beats the previously published models, creating new state-of-the-art overall performance. Extensive analyses further validate the model's effectiveness.


\bibliography{anthology,custom}

\clearpage

\appendix
\noindent \textbf{\large Appendix}
\section{The Greedy Sampling Algorithm} \label{A}

\begin{table}[h] \small
\centering
\begin{threeparttable}
\begin{tabular}{l}
\toprule
\textbf{Algorithm 1}: Greedy Sampling Algorithm                                                                                                                                                                                                                                                                                                                                                                                                                                                                                                                                                                                                                                                                                                                                                                                \\ \midrule
\begin{tabular}[c]{@{}l@{}}
\textbf{Require}: shot $K$, dev set $\textbf{X}$ of a domain $\mathcal{D}$ and its entity \\
\ \ \ \ \quad \quad \quad type set ${\mathcal{T}}$\\ 
\ \  1:\ Sort types in $\mathcal{T}$ based on their frequencies in $\textbf{X}$\\ 
\ \  2:\ $\mathcal{S}$ $\leftarrow$ $\varnothing$ \  {\textcolor[RGB]{106,153,67}{//  Initialize the support set}}\\ 
\ \ 3:\ \{Count$_i$ $\leftarrow$ 0\}\\
\ \ \ \ \ \ \ {\textcolor[RGB]{106,153,67}{//  Initialize the count of each type in $\mathcal{S}$}} \\ 
\ \ 4:\ \textbf{while} $i$ \textless \ $\mid$${\mathcal{T}}$$\mid$ \textbf{do}\\ 
\ \ 5:\ \ \quad \textbf{while} Count$_i$ \textless{} $K$ \textbf{do}\\ 
\ \ 6:\  \ \ \qquad Sample $(\mathcal{X}, \mathcal{E})$ $\in$ $\mathbf{X}$ $\  $ $ s.t.$ $\  $ ${\mathcal{T}}_i$ $\in$ $\mathcal{E}.type\tnote{1}$\\ 
\ \ \ \ \ \ \ \ \qquad {\textcolor[RGB]{106,153,67}{ //  Sample a sentence containing entities of $\mathcal{T}_i$}} \\
\ \ \ \ \ \ \ \ \qquad {\textcolor[RGB]{106,153,67}{ //  type, w/o replacement}} \\
\ \ 7:\  \quad \quad \ \ $\mathcal{S}$ $\leftarrow$ $\mathcal{S}$ $\cup$ \{$(\mathcal{X}, \mathcal{E})$\}\\ 
\ \ 8:\  \quad \quad \ \ update \{Count$_j$\} \ $\forall$\ ${\mathcal{T}}_j$ $\in$ $\mathcal{E}.type$\\ 
\ \ 9:\ \ \quad \textbf{end while}\\ 
10:\ \textbf{end while} \\
11:\ \textbf{return} $\mathcal{S}$

\\ \bottomrule
\end{tabular}
\end{tabular}
\begin{tablenotes}
\footnotesize
\item[1]  $\mathcal{E}.type$ denotes the types of entities annotated in $\mathcal{E}$
\end{tablenotes}
\end{threeparttable}
\end{table}

\section{Details of the Evaluation Task} \label{B}

\subsection{Tag Set Extension}

The Group A, B, and C split from the OntoNotes dataset are as follows.

\begin{compactitem}
\item Group A: \{\texttt{Org, Quantity, Ordinal, Norp, Work, Law}\} 
\item Group B: \{\texttt{Gpe, Cardinal, Percent, Time, Event, Language}\} 
\item Group C: \{\texttt{Person, Product, Money, Date, Loc, Fac}\}
\end{compactitem}

In this task, we evaluate our EP-Net on each group while training our model on the other two groups. In each experiment, we modify the training set by replacing all entity types in the target type set with the \texttt{None} type. Hence, these target types are no longer observed during training. We use the modified training set for model training in the \textbf{Train} step. Similarly, we modify the dev and test sets to only include entity types contained in the target type set. We use the Greedy Sampling Algorithm to sample multiple support sets from the dev set for model adaption.

\subsection{Domain Transfer}
In this task, we train our EP-Net on the standard training set of the OntoNotes dataset and evaluate our model on the standard test sets of I2B2, CoNLL, and WNUT. In addition, we sample support sets for model adaption from the standard dev sets of the above three datasets.

\subsection{Few-NERD Settings}\label{b3}

FEW-NERD \cite{dingning} is the first dataset specially constructed for few-shot NER and is one of the largest human-annotated NER datasets. It consists of 8 coarse-grained entity types and 66 fine-grained entity types. The dataset contains two sub-sets, name \textbf{Intra} and \textbf{Inter}. 
\begin{compactitem}
\item In Intra, all the fine-grained entity types belonging to the coarse-grained \texttt{People}, \texttt{MISC}, \texttt{Art}, \texttt{Product} are assigned to the training set, and all the fine-grained entity types belonging to the coarse-grained \texttt{Event}, \texttt{Building} are assigned to the dev set, and all the fine-grained entity types belonging to the coarse-grained \texttt{ORG}, \texttt{LOC} are assigned to the test set. In this dataset, the training/dev/test sets share little knowledge, making it a difficult benchmark.
\item In Inter, 60\% of the 66 fine-grained types are assigned to the training set, 20\% to the dev set, and 20\% to the test set. 
The intuition of this dataset is to explore if the coarse information will affect the prediction of new entities.
\end{compactitem}

We use the standard evaluation ($\S$\ref{3.2}) and the episode evaluation to evaluate the performance of our EP-Net. 
For the standard evaluation, we conduct experiments on Intra and Inter, respectively. We first use the training set to train our EP-Net and then sample support sets from the test set for the model adaptation and evaluate our model on the remaining test set. For the episode evaluation, we use the exact evaluation setting proposed by \cite{dingning}.

\section{Baseline Details}\label{C}
Following the established line of work \cite{yang,das,dingning}, we compare EP-Net with the following competitive models.
\begin{compactitem}
\item {Prototypical Network} (\textbf{ProtoNet}) \cite{prototypical} is a popular few-shot classification algorithm that has been adopted in most previously published token-level few-shot NER models.
\item \textbf{ProtoNet+P\&D}  \cite{hou} uses pair-wise embedding and collapsed dependency transfer mechanism in the token-level Prototypical Network, tackling challenges of similarity computation and transferring estimated label dependency across domains. 
\item \textbf{NNShot} \cite{yang} is a simple token-level nearest neighbor classification model. It simply computes a similarity score between a token in the query example and all tokens in the support set.
\item \textbf{StructShot} \cite{yang} combines NNShot and Viterbi decoder and uses estimated label dependency across domains by first learning abstract label dependency and then distributing it evenly to target domains.
\item \textbf{CONTaiNER} \cite{das} introduces Contrast Learning to the StructShot. It models Gaussian embedding and optimizes inter token distribution distance, which aims to decrease the distance of token embeddings of similar entities while increasing the distance for dissimilar ones.
\end{compactitem}

For a fair comparison, we use the results of the ProtoNet, ProtoNet+P\&D, NNShot, and StructShot reported in \cite{yang}, and the results of CONTaiNER reported in \cite{das}. 

In addition, we run the ProtoNet, {ProtoNet+P\&D}, {NNShot}, and {StructShot} on Few-NERD using the standard evaluation setup ($\S$\ref{3.2}, \ref{b3}).



\section{Performance against Prototype Distance Threshold ($\tau$)} \label{D}
\begin{figure}[h]
\centering
\includegraphics[width=0.42\textwidth]{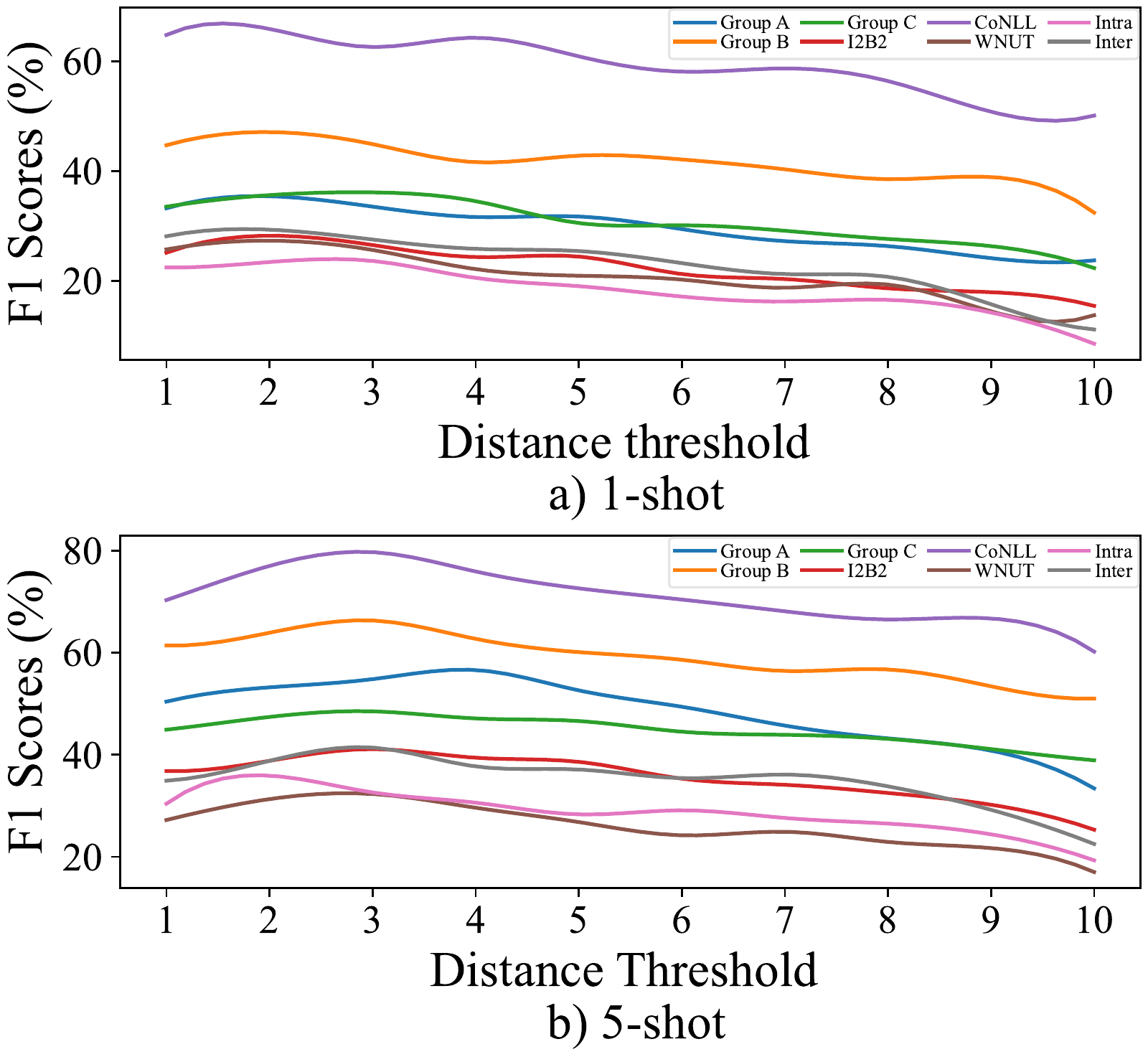} 
\caption{Performance comparisons of different prototype distance threshold ($\tau$) values in 1- and 5-shot experiments.} 
\label{figure7}
\end{figure}

We conduct 1- and 5-shot experiments to explore the performance against different $\tau$ values. Since validation sets are unavailable in the few-shot scenario, we randomly sample 20\% of the query sets for the explorations. We report the results in Figure \ref{figure7}, where we set the $\tau$ value from 1 to 10, respectively. We can observe that: (1) The F1 scores generally first increase and then decrease when the $\tau$ value consistently increases. (2) Except for the Group C and Intra, our EP-Net performs the best in the 1-shot experiments when setting the $\tau$ to 2. (3) Except for the Group A and Intra, our EP-Net performs the best in the 5-shot experiments when setting the $\tau$ to 3. 

The above results validate our argument that the prototypes should be distributed in an appropriate-sized embedding space, neither too large nor too small ($\S$\ref{4.1}).
For simplicity, we set the $\tau$ to 2 and 3 in all the other 1- and 5-shot experiments, respectively.

\section{Episode Evaluation on Few-NERD} \label{E}
We evaluate our EP-Net on Few-NERD with the episode evaluation setting and compare our model with previous state-of-the-art models, including ProtoBERT \cite{dingning}, NNShot, StructShot, CONTaiNER, and ESD \cite{wangpeiyi}.  We would like to mention that the ESD is a concurrent span-based few-shot NER model to ours.

\begin{table*}[h]\small
\centering
\begin{tabular}{lcccccl}
\toprule
\multirow{2}{*}{Model} & \multicolumn{2}{c}{1$\sim$2-shot} &  & \multicolumn{2}{c}{5$\sim$10-shot} & \multicolumn{1}{c}{\multirow{2}{*}{Avg.}} \\ \cmidrule{2-3} \cmidrule{5-6}
                       & 5 way           & 10 way          &  & 5 way            & 10 way          & \multicolumn{1}{c}{}                      \\ \midrule
ProtoBERT              & 23.45$\pm$0.92      & 19.76$\pm$0.59      &  & 41.93$\pm$0.55       & 34.61$\pm$0.59      &29.94                                           \\
NNShot                 & 31.01$\pm$1.21      & 21.88$\pm$0.23      &  & 35.74$\pm$2.36       & 27.67$\pm$1.06      &29.08                                           \\
StructShot             & 35.92$\pm$0.69      & 25.38$\pm$0.84      &  & 38.83$\pm$1.72       & 26.39$\pm$2.59      &31.63                                           \\
ESD                    & 41.44$\pm$1.16      & 32.29$\pm$1.10      &  & 50.68$\pm$0.94       & 42.92$\pm$0.75      &41.83                                           \\ 
CONTaiNER              & 40.43           & 33.84           &  & 53.70            & \textbf{47.49}           &                                          43.87 \\
\midrule
EP-Net (Ours)          & \textbf{43.36$\pm$0.99}               &  \textbf{36.41$\pm$1.03}                &  & \textbf{58.85$\pm$1.12}                 &46.40$\pm$0.87                 & \textbf{46.26}                                          \\ \bottomrule
\end{tabular}
\caption{Episode evaluation results (F1 scores) on the Intra dataset of Few-NERD. We report the mean and standard deviations of F1 scores.}\label{table4}
\end{table*}

\begin{table*}[h]\small
\centering
\begin{tabular}{lcccccl}
\toprule
\multirow{2}{*}{Model} & \multicolumn{2}{c}{1$\sim$2-shot} &  & \multicolumn{2}{c}{5$\sim$10-shot} & \multicolumn{1}{c}{\multirow{2}{*}{Avg.}} \\ \cmidrule{2-3} \cmidrule{5-6}
                       & 5 way           & 10 way          &  & 5 way            & 10 way          & \multicolumn{1}{c}{}                      \\ \midrule
ProtoBERT              & 44.44$\pm$0.11      & 39.09$\pm$0.87      &  & 58.80$\pm$1.42       & 53.97$\pm$0.38      &49.08                                           \\
NNShot                 & 54.29$\pm$0.40      & 46.98$\pm$1.96      &  & 50.56$\pm$3.33       & 50.00$\pm$0.36      &50.46                                           \\
StructShot             & 57.33$\pm$0.53      & 49.46$\pm$0.53      &  & 57.16$\pm$2.09       & 49.39$\pm$1.77      & 53.34                                          \\
CONTaiNER              & 55.95           & 48.35           &  & 61.83            & 57.12           &                                          55.81 \\
ESD                    & \textbf{66.46$\pm$0.49}      & \textbf{59.95$\pm$0.69}      &  & \textbf{74.14$\pm$0.80}       & \textbf{67.91$\pm$1.41}      &\textbf{66.12}                                           \\ \midrule
EP-Net (Ours)          &62.49$\pm$0.36                 &54.39$\pm$0.78                 &  &65.24$\pm$                 0.64 &  62.37$\pm$1.27                  &61.12                                           \\ \bottomrule
\end{tabular}
\caption{Episode evaluation results (F1 scores) on the Inter dataset of Few-NERD. We report the mean and standard deviations of F1 scores.}\label{table5}
\end{table*}

We report the results in Table \ref{table4} and Table \ref{table5}, where we take the results of ProtoBERT, NNShot, and StructShot reported in \cite{dingning}, and the results of CONTaiNER and ESD reported in their original papers.
We can see that:
\begin{compactitem}
\item On the Intra, our EP-Net consistently outperforms the best baseline (i.e., CONTaiNER) in terms of the Avg. metric, bringing +2.39\% F1 gains. In addition, our EP-Net surpasses the concurrent ESD by +4.43\% F1 scores. 
\item On the Inter, our EP-Net is inferior to ESD by a large margin (5.0\%) in terms of the Avg. metric. However, our model consistently outperforms the other baselines, delivering up to +5.31\% F1 scores compared to the CONTaiNER.
\item Both our EP-Net and CONTaiNER outperform ESD in 1-shot experiments, but they are inferior to ESD in 5-shot experiments. 
\end{compactitem}

The above results demonstrate the effectiveness of the proposed EP-Net. And compared to ESD, our model is more efficient in the few-shot scenario when entities share less coarse-grained information (the Intra).\footnote{As shown in Appendix \ref{b3}, entities in the Intra share little coarse-grained information, but the Inter is designed to allow entities sharing the coarse-grained information.}

Compared to our simple concatenation method (Eq.\ref{6b}) to obtain span representations, ESD proposes to use Inter Span Attention (ISA) and Cross Span Attention (CSA) to enhance the span representations. We believe that the ISA and CSA enable ESD to encode the shared coarse-grained information into span representations sufficiently, which helps ESD obtain the current state-of-the-art performance on the Inter dataset.

\section{CP-Net}\label{F}
We propose the CP-Net as a comparable model to our EP-Net. CP-Net is also an entity-level prototypical network, but it uses conventional prototypes obtained by averaging the embeddings of type's examples. Similar to EP-Net, CP-Net also uses the BERT model as an embedding generator. 
In addition, it uses the sampling strategy discussed in $\S$\ref{5.3} to randomly sample \texttt{None} spans. 
CP-Net consists of two steps, namely \textbf{Train} and \textbf{Recognize}. 

In the Train step, we train CP-Net with the source domain data. To be specific, we obtain the entity-level prototypes by averaging the embeddings of type's examples in the training set. Moreover, we obtain span representations with the same method of EP-Net (Eq.\ref{eq3}-\ref{eq7}), as well as the method to calculate span-prototype similarity (Eq.\ref{eq8}-\ref{eq9}). During the model training, we use the training loss $\mathcal{L}_s$ (Eq.\ref{eq10b}) to fine-tune the BERT model. 

In the Recognize step, we use the fine-tuned BERT model as the embedding generator and obtain the entity-level prototypes by averaging the embeddings of each type's examples in the support sets. Then we obtain the type of each span according to the best similarity between the span and the prototypes.  

The CP-Net differs from our EP-Net in the following two ways.
\begin{compactitem}
\item CP-Net uses conventional prototypes, and it does not train these prototypes during the model training. By contrast, our EP-Net trains prototypes from scratch with the distance based loss $\mathcal{L}_d$ (Eq.\ref{2c})
\item CP-Net does not contain a domain adaption procedure, and it solely uses the support sets for similarity calculation. By contrast, our EP-Net contains a \textbf{Adapt} step for domain adaption and it uses the support sets for not only the similarity calculation but also the domain adaption.
\end{compactitem}

\end{document}